\pdfoutput=1

\documentclass[11pt]{article}

\usepackage[final]{coling}

\usepackage{times}
\usepackage{latexsym}

\usepackage[T1]{fontenc}

\usepackage[utf8]{inputenc}

\usepackage{microtype}

\usepackage{inconsolata}

\usepackage{amsmath}
\usepackage{amssymb}

\usepackage{graphicx}
\usepackage{subfigure}

\usepackage{array} 
\usepackage{tabularray}
\usepackage{booktabs}
\usepackage{longtable}
\usepackage{multirow}
\usepackage{multicol}
\usepackage{bigstrut}
\usepackage{makecell}

\usepackage{algorithm}
\usepackage{algorithmic}

\usepackage{hyperref}

\DeclareUnicodeCharacter{0308}{\"}

\title{BiLD: Bi-directional Logits Difference Loss for \\ Large Language Model Distillation}

\author{
 \textbf{Minchong Li\textsuperscript{1,2}}
 \qquad
 \textbf{Feng Zhou\textsuperscript{2}}
 \qquad
 \textbf{Xiaohui Song\textsuperscript{2} \thanks{Corresponding author.}}
\\
 \textsuperscript{1}KTH Royal Institute of Technology, Stockholm, Sweden
\\
 \textsuperscript{2}OPPO AI Center, Beijing, China
\\
 \texttt{mincli@kth.se}
\\
 \texttt{\{zhoufeng1, songxiaohui\}@oppo.com}
}

\begin{document}
\maketitle
\begin{abstract}
In recent years, large language models (LLMs) have shown exceptional capabilities across various natural language processing (NLP) tasks. However, such impressive performance often comes at the cost of an increased parameter size, posing significant challenges for widespread deployment. To address this issue, knowledge distillation (KD) provides a solution by transferring knowledge from a large teacher model to a smaller student model. In this paper, we explore the task-specific distillation of LLMs at the logit level. Our investigation reveals that the logits of fine-tuned LLMs exhibit a more pronounced long-tail distribution compared to those of vision models, with hidden "noise" in the long tail affecting distillation performance. Furthermore, existing logits distillation methods often struggle to effectively utilize the internal ranking information from the logits. To address these, we propose the \textbf{Bi}-directional \textbf{L}ogits \textbf{D}ifference (BiLD) loss. The BiLD loss filters out the long-tail noise by utilizing top-$k$ teacher and student logits, and leverages the internal logits ranking information by constructing logits differences. To evaluate BiLD loss, we perform comprehensive experiments on 13 datasets with two types of LLMs. Our results show that the BiLD loss, with only the top-\textbf{8} logits, outperforms supervised fine-tuning (SFT), vanilla KL loss, and five other distillation methods from both NLP and CV fields. Code
is available at \url{https://github.com/fpcsong/BiLD}.
\end{abstract}

\section{Introduction}
\label{sec:intro}

\begin{figure*}[ht]
\centering
\includegraphics[width=1\textwidth]{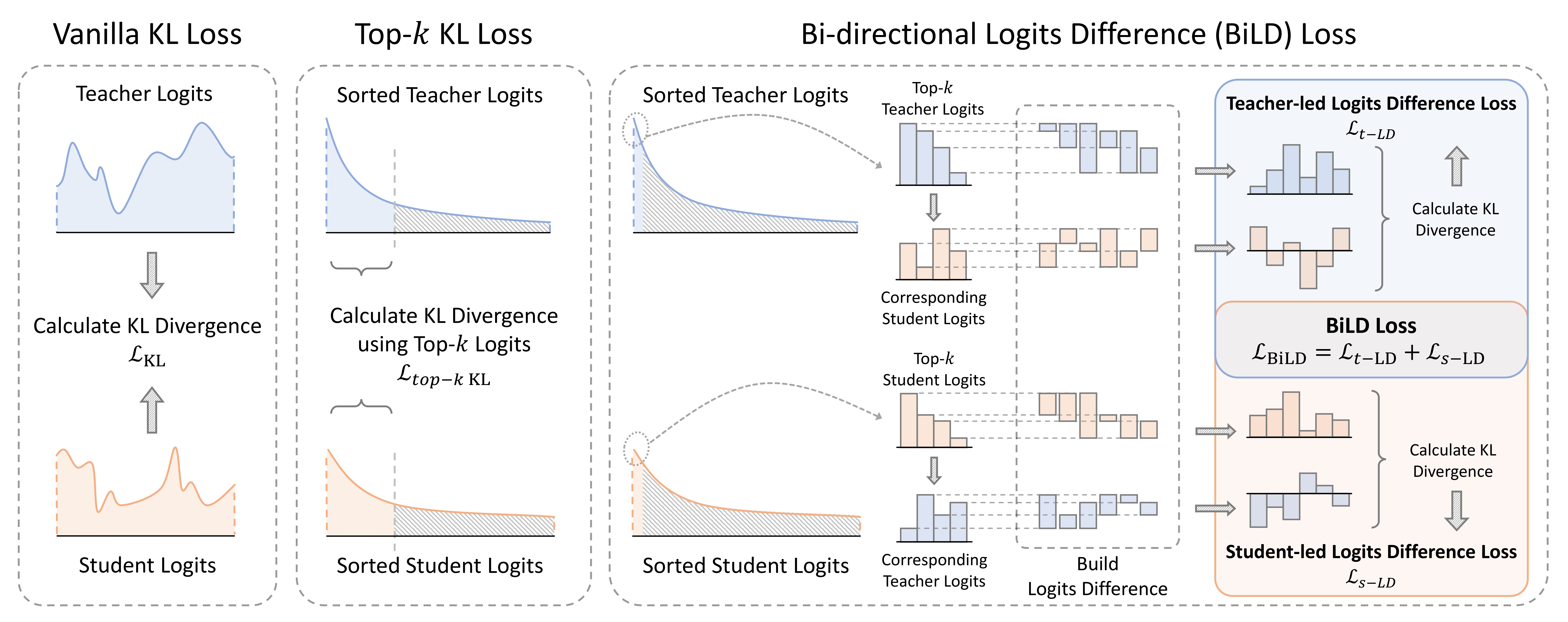}
\caption{An illustration of vanilla KL, top-$k$ KL and our BiLD loss. The vanilla KL loss directly calculates the KL divergence between teacher and student logits, whereas the top-$k$ KL loss uses clipped logits instead of the full logits. In contrast to these methods, our BiLD loss computes KL divergence based on reconstructed logits differences. The logits difference is derived by calculating the pairwise differences between logit values. We construct two groups of logits differences and compute the KL divergence within each group as a loss: the top-$k$ teacher logits and their corresponding student logits are used to calculate the teacher-led logits difference ($t$-LD) loss, while the top-$k$ student logits and their corresponding teacher logits are used to calculate the student-led logits difference ($s$-LD) loss. The BiLD loss is the sum of these two losses.}
\label{fig:bild}
\end{figure*}

The last few years have witnessed large language models (LLMs) risen to prominence, demonstrating remarkable proficiency in natural language understanding and generation. However, these capabilities come at the cost of an ever-increasing number of parameters. Due to constraints on computational resources, LLMs' formidable size limits their democratization and widespread deployment. Knowledge distillation (KD) , as a classic model compression method\citep{hinton2015distilling}, provides a solution for reducing model size while preserving performance. KD transfers knowledge from a large teacher model to a smaller student model, enhancing the latter's performance and enabling its lightweight deployment.

As an important branch of KD, logits distillation has gained popularity due to its straightforward application. The goal of logits distillation is to minimize the Kullback-Leibler (KL) divergence between the teacher and student logits. A significant portion of research on logits distillation has focused on vision models \citep{zhao2022decoupled, yang2023multi, chi2023normkd, Sun2024Logit}. However, the application of these methods to distill LLMs has yet to be thoroughly explored due to potential differences in structure, data distribution, and output space between vision and language models.

For LLMs, research on logits distillation is still emerging, with methods such as reverse KL \citep{tu2020engine, lee2023self, gu2023minillm} and those based on optimal transport metrics \citep{cui2024sinkhorn}. However, in practical applications, the former suffers from the "mode-seeking" problem \citep{chan2022greedification, li2023mode}, while the latter is computationally too complex for open-source LLMs with billions of parameters.

In this paper, we investigate the characteristics of logits in LLMs. Compared to the limited output space of vision models, LLMs' output space comprises sequences of discrete tokens of potentially infinite length, making LLM logits significantly more complex. Furthermore, LLM logits exhibit a noticeable long-tail distribution, indicating a substantial portion of "noise" beyond a small amount of key knowledge. We also observe that in LLM text generation, common strategies like top-$k$ sampling and top-$p$ sampling are influenced by the internal ranking of logits when selecting output tokens. However, existing logits distillation methods often struggle to exploit this latent ranking information \citep{Sun2024Logit}.

Motivated by the characteristics of logits and the underutilization of their internal ranking, we propose a novel loss function, called Bi-directional Logits Difference (BiLD) loss, for task-specific LLM distillation. BiLD loss emphasizes reducing long-tail noise and explicitly utilizes the ranking information in logits. It computes KL divergence based on reconstructed logits differences, which are obtained by calculating the internal pairwise differences of values from top-$k$ teacher (student) logits and the corresponding student (teacher) logits. Our experiments show that BiLD loss, using only the top-8 logits, achieves state-of-the-art (SOTA) results across various NLP tasks. Our code is available at \href{https://github.com/fpcsong/BiLD}{https://github.com/fpcsong/BiLD}.

To conclude, we make the following contributions:

\begin{itemize}
    \item We investigate the characteristics of LLMs' logits, discussing their internal distribution and the significance of logits internal ranking.
    \item We propose the Bi-directional Logits Difference (BiLD) loss for logits distillation of LLMs. BiLD filters out "noise" in logits' long-tail distribution while leveraging logits ranking information to enhance performance. Our method can serve as an alternative to the vanilla KL loss in existing LLM distillation methods.
    \item To demonstrate the effectiveness of BiLD loss, we conduct comprehensive experiments on 13 NLP datasets using two open-source LLMs, BLOOM \citep{bigscience_workshop_2022} and Qwen1.5 \citep{qwen}. We evaluate various logits distillation methods from both CV and NLP domains. Experimental results show that our BiLD loss outperforms SFT, vanilla KL loss and five other methods using only the top-8 logits. Furthermore, our comparison of teacher and student logits shows that BiLD loss promotes better imitation of teacher's primary behavior at the logit level.
\end{itemize}

\section{Related Works}
\label{sec:related works}

\subsection{Logits Distillation}
\label{sec:logits distillation}

One representative approach of knowledge distillation is logits distillation, which transfers knowledge by minimizing the divergence of output logits \citep{jin2023multi}. For vision models, there has been substantial research on logits distillation. Approaches like DKD \citep{zhao2022decoupled} and NKD \citep{yang2023knowledge} decouple the target and non-target components of logits, applying weighting or regularization. NormKD \citep{chi2023normkd} dynamically customizes temperatures during the distillation process. However, the differences in structure, data, and output space between vision models and LLMs make it challenging to directly apply these methods to LLMs.

Recent research has introduced several logit distillation methods specifically for LLMs. Reverse KL (RKL) \citep{tu2020engine, gu2023minillm} has been used to mitigate the "mode-averaging" problem; however, it sometimes leads the student model towards "mode-seeking" behavior. DistiLLM \citep{ko2024distillm} proposes mixing the logits distributions of the teacher and the student, but this introduces additional hyperparameters, increasing its complexity in practical applications. SinKD \citep{cui2024sinkhorn} replaces KL divergence with Sinkhorn Distance, but its computational demands pose challenges when applied to larger models.

Our work continues the paradigm of reducing the divergence of logits. However, unlike previous approaches, we calculate the divergence of logits differences instead of logits themselves. Our method focuses the model on the key knowledge in the teacher logits without introducing excessive hyperparameters.

\subsection{Other Distillation Methods for LLMs}
\label{sec:distilllation llms}

Previous works on distillation for LLMs extend beyond logits-based methods, primarily falling into two categories: white-box and black-box approaches \citep{yuan2024llm}. White-box distillation \citep{gu2023minillm, liang2023less, agarwal2024policy} leverages the teacher's internal representations and hidden states to facilitate knowledge transfer. However, these methods often rely on structural similarities between the teacher and student models. In contrast, black-box distillation only permits the student to access the teacher's outputs. Current research in black-box distillation mainly focuses on learning from the teacher's output texts \citep{sahu2023promptmix, fu2023specializing, li2023symbolic}. While BiLD can be classified as black-box distillation, it serves as an alternative to the vanilla KL divergence loss and can be easily integrated with white-box distillation methods.

\section{Methods}
\label{sec:methods}

\subsection{Brief Review of Logits Distillation}
\label{sec:methods review}

Logits distillation calculates the divergence between the teacher's and student's logits as the optimization target. Consider a teacher model $t$ and a student model $s$, both with a vocabulary size $N$. During the process of single token prediction, the teacher logits $\textbf{z}^t$ and student logits $\textbf{z}^s$ at a certain position can be represented as: 

\begin{equation}
\label{eq:kl_logits}
\begin{split}
&\textbf{z}^t = \left[ z^t_1, z^t_2, \cdots , z^t_N \right] \in \mathbb{R}^{1\times N},\\
&\textbf{z}^s = \left[ z^s_1, z^s_2, \cdots , z^s_N \right] \in \mathbb{R}^{1\times N}.
\end{split}
\end{equation}

Logits are the raw outputs of language models and cannot be directly used to calculate the loss. We process the logits into probabilities $\textbf{p}^t$ and $\textbf{p}^s$, where an element $p_i$ from $\textbf{p}^t$ or $\textbf{p}^s$ represents the probability of the token at the $i$-th position being sampled as the output:

\begin{equation}
\label{eq:kl_probs}
\begin{split}
&\textbf{p}^t = \frac{\textrm{exp}(\textbf{z}^t / T)}{\sum\nolimits_{N}^{i=1} \textrm{exp} (z^t_i / T)} \in \mathbb{R}^{1\times N},\\
&\textbf{p}^s = \frac{\textrm{exp}(\textbf{z}^s / T)}{\sum\nolimits_{N}^{i=1} \textrm{exp} (z^s_i / T)} \in \mathbb{R}^{1\times N},
\end{split}
\end{equation}

\noindent where $T$ is the temperature during normalization. The vanilla KL divergence loss is defined as: 

\begin{equation}
\label{eq:kl_formula}
\mathcal{L} _{\rm KL}= D_{\rm KL} \left [ \textbf{p}^t \parallel \textbf{p}^s \right ].
\end{equation}

By aligning the student's logits with that of the teacher using vanilla KL loss, the student can imitate the teacher's performance at the logit level, thereby facilitating knowledge transfer.

\subsection{The Characteristics of LLMs' Logits}
\label{sec:characteristics of logits}

Compared to vision models, LLMs have an output space consisting of infinitely long sequences of tokens, which makes their logits more complex. To compare the logit characteristics of vision models and LLMs, we conduct a toy experiment using ResNet-101 \citep{he2016deep} and Qwen-4B \citep{qwen}. In this experiment, we randomly select five images and five sets of instructions from our test data as inputs for the vision and language models (details about images and instructions are provided in Appendix~\ref{app:toy_case}). We use kurtosis to measure the extremity of logits' long-tail distribution and calculate the proportion of top-$k$ logit values. The experimental results are reported in Table~\ref{tab:toy_results}. The kurtosis of text logits is 2-3 orders of magnitude higher than that of image logits, suggesting that text logits are much "sharper" than image logits. Given that text logits are much longer than image logits, the proportion of top-$k$ logit values also indicates that text logit values are more peaked than those of image logits.

\begin{table*}[!ht]
  \centering
    \begin{tabular}{ccccccc}
    \Xhline{1pt}
    \multirow{2}[4]{*}{Input Image / Text} & \multirow{2}[4]{*}{Model} & \multirow{2}[4]{*}{Kurtosis} & \multicolumn{4}{c}{Top-$k$ logits percentage (\%)} \bigstrut\\
\cline{4-7}          &       &       & $k$=8   & $k$=64  & $k$=512 & $k$=1024 \bigstrut\\
    \hline
    \hyperref[fig:toy_images]{cat.jpg} & \multirow{5}[2]{*}{ResNet-101} & 975   & 99.540\% & 99.642\% & 99.993\% & \textbackslash{} \bigstrut[t]\\
    \hyperref[fig:toy_images]{dogs.jpg} &       & 782   & 93.977\% & 98.433\% & 99.882\% & \textbackslash{} \\
    \hyperref[fig:toy_images]{lioness.jpg} &       & 995   & 99.904\% & 99.973\% & 99.999\% & \textbackslash{} \\
    \hyperref[fig:toy_images]{mushroom.jpg} &       & 914   & 99.756\% & 99.968\% & 99.998\% & \textbackslash{} \\
    \hyperref[fig:toy_images]{hat.jpg} &       & 906   & 83.982\% & 93.643\% & 99.646\% & \textbackslash{} \bigstrut[b]\\
    \hline
    \hyperref[app:ins_1]{Instruction 1}  & \multirow{5}[2]{*}{Qwen-4B} & 135404 & 99.991\% & 99.996\% & 99.997\% & 99.998\% \bigstrut[t]\\
    \hyperref[app:ins_2]{Instruction 2} &       & 46163 & 99.998\% & 99.998\% & 99.998\% & 99.998\% \\
    \hyperref[app:ins_3]{Instruction 3} &       & 79604 & 99.982\% & 99.990\% & 99.993\% & 99.994\% \\
    \hyperref[app:ins_4]{Instruction 4} &       & 50719 & 99.528\% & 99.604\% & 99.634\% & 99.651\% \\
    \hyperref[app:ins_5]{Instruction 5} &       & 116329 & 94.778\% & 94.826\% & 94.977\% & 95.081\% \bigstrut[b]\\
    \Xhline{1pt}
    \end{tabular}%
  \caption{The kurtosis and top-$k$ proportion of image logits and text logits.}
  \label{tab:toy_results}%
\end{table*}%

Moreover, previous logits distillation methods have not fully utilized the internal rank information of logits \citep{huang2022knowledge, Sun2024Logit}, even though this ranking information significantly affects LLMs' generation performance. When LLMs generate text, two sampling strategies, top-$k$ sampling and top-$p$ sampling, are commonly used to control the diversity of the generated content. Top-$k$ sampling controls the maximum length of the candidate tokens list, while top-$p$ sampling filters tokens according to cumulative probability. The ranking of logit values impacts the selection process in both strategies, as higher-ranked tokens are more likely to be selected as candidates. Therefore, maintaining rank consistency will better assist the student in imitating the teacher's generating patterns.

\subsection{Bi-directional Logits Difference Loss}
\label{sec:bild}

The Bi-directional Logits Difference (BiLD) loss is a novel optimization target for task-specific LLM distillation. It filters out the "noise" in the long-tail distribution of LLMs' logits and constructs bi-directional differences that reflect the internal ranking of logits. Our goal is not for the student logits to fully match the teacher's but for the student to effectively learn the key knowledge represented in the non-long-tail part. The detailed process of BiLD is shown in Figure~\ref{fig:bild}.

\subsubsection{Formal Definition}
\label{sec:formal_definition}

The BiLD loss consists of two components: the teacher-led logits difference ($t$-LD) loss and the student-led logits difference ($s$-LD) loss. Due to the similarity between the two components, we explain the process by calculating the $t$-LD loss. First, we select the top-$k$ teacher logits and sort them in descending order to build the teacher-led logits $\textbf{z}^{t}_{\rm led}$:

\begin{equation}
\label{eq:tld_select_t}
\textbf{z}^{t}_{\rm led} = \left[ z^t_{i_1}, z^t_{i_2}, \cdots , z^t_{i_k} \right] \in \mathbb{R}^{1\times k},
\end{equation}

\noindent where the elements of $\textbf{z}^{t}_{\rm led}$ satisfy $z^t_{i_1} \ge z^t_{i_2} \ge \cdots \ge z^t_{i_k}$. Then, we create the corresponding student logits $\textbf{z}^{s}_{\rm cor}$ by selecting the student logit values at the corresponding positions $\left[i_1, i_2, \cdots, i_k \right]$:

\begin{equation}
\label{eq:tld_select_s}
\textbf{z}^{s}_{\rm cor} = \left[ z^s_{i_1}, z^s_{i_2}, \cdots , z^s_{i_k} \right] \in \mathbb{R}^{1\times k}.
\end{equation}

Next, we build the logits differences $\textbf{d}^t_{\rm led}$ and $\textbf{d}^s_{\rm cor}$ by calculating the internal pairwise value differences of $\textbf{z}^{t}_{\rm led}$ and $\textbf{z}^{s}_{\rm cor}$ respectively:

\begin{equation}
\label{eq:tld_build_diff}
\begin{split}
& \textbf{d}^t_{\rm led} = \left[ z^t_{i_m} - z^t_{i_n} \mid 1 \leq m < n \leq k \right],\\
& \textbf{d}^s_{\rm cor} = \left[ z^s_{i_m} - z^s_{i_n} \mid 1 \leq m < n \leq k \right],
\end{split}
\end{equation}

\noindent where both $\textbf{d}^t_{\rm led}$ and $\textbf{d}^s_{\rm cor}$ $\in \mathbb{R}^{1 \times \frac{k(k-1)}{2}}$. Then we normalize $\textbf{d}^t_{\rm led}$ and $\textbf{d}^s_{\rm cor}$ into probabilities:

\begin{equation}
\label{eq:tld_probs}
\begin{split}
&\textbf{p}^t_{\rm led} = \frac{\textrm{exp}(\textbf{z}^t_{\rm led} / T)}{\sum\nolimits_{i=1}^{\frac{k(k-1)}{2}} \textrm{exp} (z^t_{{\rm led}, i} / T)},\\
&\textbf{p}^s_{\rm cor} = \frac{\textrm{exp}(\textbf{z}^s_{\rm cor} / T)}{\sum\nolimits_{i=1}^{\frac{k(k-1)}{2}} \textrm{exp} (z^s_{{\rm cor}, i} / T)}.
\end{split}
\end{equation}

To obtain the teacher-led logits difference loss $\mathcal{L} _{t- \rm LD}$, we calculate the KL divergence between $\textbf{p}^t_{\rm led}$ and $\textbf{p}^s_{\rm cor}$:

\begin{equation}
\label{eq:tld_formula}
\mathcal{L} _{t- \rm LD} = D_{\rm KL} \left [ \textbf{p}^t_{\rm led} \parallel \textbf{p}^s_{\rm cor} \right ].
\end{equation}

The calculation of the $s$-LD loss is similar to that of the $t$-LD loss. The key difference is that the $s$-LD loss selects the top-$k$ student logits $\textbf{z}^s_{\rm led}$ and extracts the corresponding teacher logits $\textbf{z}^t_{\rm cor}$. Based on these, we can sequentially calculate the logits differences $\textbf{d}^s_{\rm led}$ and $\textbf{d}^t_{\rm cor}$ as well as the probabilities $\textbf{p}^s_{\rm led}$ and $\textbf{p}^t_{\rm cor}$. The $s$-LD loss can be represented as:

\begin{equation}
\label{eq:sld_formula}
\mathcal{L} _{s- \rm LD} = D_{\rm KL} \left [ \textbf{p}^t_{\rm cor} \parallel \textbf{p}^s_{\rm led} \right ].
\end{equation}

Finally, we obtain the BiLD loss:

\begin{equation}
\label{eq:bild_formula}
\mathcal{L} _{\rm BiLD} = \mathcal{L}_{t- \rm LD} + \mathcal{L}_{s- \rm LD}.
\end{equation}

To aid comprehension, we outline the calculation process of the BiLD loss in Algorithm \ref{alg:bild}. 

\begin{algorithm}[!ht]
	\caption{Calculation of BiLD Loss} 
	\label{alg:bild} 
	\renewcommand{\algorithmicrequire}{\textbf{Input:}}
	\renewcommand{\algorithmicensure}{\textbf{Output:}}
	\begin{algorithmic}[1]
		\REQUIRE teacher logits $\textbf{z}^t$, student logits $\textbf{z}^s$, temperature $T$, hyperparameter $k$ that controls the number of clipped logits
        \ENSURE the BiLD loss $\mathcal{L}_{\rm BiLD}$ 
        \STATE select top-$k$ teacher logits $\textbf{z}^t_{\rm led}$ (Equation~\ref{eq:tld_select_t})
        \STATE select corresponding student logits $\textbf{z}^s_{\rm cor}$ (Equation~\ref{eq:tld_select_s})
        \STATE build the teacher and student logits differences $\textbf{d}^t_{\rm led}$ and $\textbf{d}^s_{\rm cor}$ (Equation~\ref{eq:tld_build_diff})
        \STATE normalize differences to probabilities $\textbf{p}^t_{\rm led}$ and $\textbf{p}^s_{\rm cor}$ (Equation~\ref{eq:tld_probs})
        \STATE calculate the teacher-led logits difference loss $\mathcal{L}_{t-{\rm LD}}$ (Equation~\ref{eq:tld_formula})
        \STATE calculate $\mathcal{L}_{s-{\rm LD}}$ (Equation~\ref{eq:sld_formula}), generally following steps 1-5
        \STATE sum $\mathcal{L}_{t-{\rm LD}}$ and $\mathcal{L}_{s-{\rm LD}}$ to obtain $\mathcal{L} _{\rm BiLD}$ (Equation~\ref{eq:bild_formula})
	\end{algorithmic} 
\end{algorithm}

\subsubsection{Explanation about the Utilization of Logits Ranking}

The calculation of the logits difference (Equation~\ref{eq:tld_build_diff}) ensures that the student model learns the ranking information embedded in the teacher logits. We demonstrate this through an example based on the $t$-LD loss calculation. Since $\textbf{z}^t_{\rm led}$ satisfies $z^t_{i_1} \ge z^t_{i_2} \ge \cdots \ge z^t_{i_k}$, it is guaranteed that every element in the teacher-led logits difference $\textbf{d}^t_{\rm led}$ is non-negative. For the corresponding student logits difference $\textbf{d}^s_{\rm cor}$, consider an element $d^s = z^s_{i_{m}} - z^s_{i_{n}}$. If $m > n$, then $d^s < 0$. In this case, the order $z^s_{i_{m}} < z^s_{i_{n}}$ is inconsistent with the order in the teacher logits $z^t_{i_{m}} > z^t_{i_{n}}$. Therefore, the sign of the elements in the corresponding logits difference $\textbf{d}^s_{\rm cor}$ reflects whether the ranking of the teacher and student logits value pairs is consistent. When calculating $\mathcal{L} _{s- {\rm LD}}$, this acts as a constraint, promoting the student logits to align their ranking order with the teacher logits. 

\section{Experiments}
\label{sec:experiments}

\subsection{Datasets}
\label{sec:datasets}

We evaluate our BiLD loss on 13 NLP datasets: (1) 8 datasets from the SuperGLUE benchmark \citep{wang2019superglue}, including BoolQ \citep{clark2019boolq}, CB \citep{de2019commitmentbank}, COPA \citep{roemmele2011choice}, MultiRC \citep{khashabi2018looking}, ReCoRD \citep{zhang2018record}, RTE \citep{giampiccolo2007third}, WiC \citep{pilehvar2018wic} and WSC \citep{levesque2012winograd}; (2) 5 extra datasets used in previous works about model compression \citep{ma2023llm, egiazarian2024extreme}, including: Arc-C, Arc-E \citep{clark2018think}, HellaSwag \citep{zellers2019hellaswag}, PIQA \citep{bisk2020piqa} and WinoGrande \citep{sakaguchi2021winogrande}. We observe that these datasets vary significantly in size (the visualization of dataset sizes is presented in Appendix~\ref{app:details_about_datasets}). Using small datasets alone for SFT and distillation would result in severe overfitting. To prevent unreliable experimental results, we use these datasets collectively for SFT and distillation, then evaluate each separately.

\subsection{Baselines}
\label{sec:baselines}

We compare BiLD loss with seven baselines: (1) supervised fine-tuning (SFT), where all parameters are adjusted during adaptation to downstream tasks; (2) vanilla KL loss; (3) vanilla KL loss with only top-$k$ logits (short as top-$k$ KL), to demonstrate the performance improvements from noise filtering; (4) three logits distillation methods for vision models, including DKD \citep{zhao2022decoupled}, NKD \citep{yang2023knowledge}, and NormKD \citep{chi2023normkd}; (5) Reverse KL loss (RKL) used in MiniLLM \citep{gu2023minillm}, which has been proven to enhance distillation performance on LLMs.

\begin{table*}[!ht]
  \renewcommand\arraystretch{1.4}
  \centering
  \resizebox{\linewidth}{!}{%
    \begin{tabular}{crcccccccccccccc}
    \Xhline{2pt}
    \multirow{2}[2]{*}{Model} & \multicolumn{1}{c}{\multirow{2}[2]{*}{Method}} & Arc-C & Arc-E & boolQ & CB & COPA & HellaSwag & MultiRC & PIQA & ReCoRD & RTE & WiC & WinoGrande & WSC & \multirow{2}[2]{*}{Avg.} \bigstrut[t]\\
          &       & (Acc.) & (Acc.) & (Acc.) & (Acc.) & (Acc.) & (Acc.) & (F1a/EM) & (Acc.) & (F1/Acc.) & (Acc.) & (Acc.) & (Acc.) & (Acc.) &  \bigstrut[b]\\
    \hline
    BLOOM-7B & Teacher & 50.84 & 68.95 & 85.26 & 89.29 & 81.00 & 76.08 & 81.36/40.82 & 74.92 & 79.87/78.50 & 83.03 & 72.41 & 71.51 & 65.38 & 72.15 \bigstrut\\
    \hline
    \multirow{8}[6]{*}{BLOOM-3B} & SFT   & 44.15 & 61.75 & 84.04 & 87.50 & 67.00 & 57.00 & 77.09/36.20 & 70.84 & 76.05/74.59 & 78.34 & 69.75 & 69.69 & 64.42 & 66.56 \bigstrut[t]\\
          & Vanilla KL & 49.50 & 68.07 & 84.50 & 87.50 & 76.00 & 72.60 & 78.89/36.52 & 74.27 & 79.81/78.32 & 81.59 & 71.94 & 70.96 & \textbf{74.04} & 71.21 \\
          & RKL   & \textbf{50.50} & 68.42 & 84.62 & 87.50 & \textbf{80.00} & 72.20 & 78.95/36.41 & 74.48 & 79.63/78.13 & 82.31 & 72.57 & 71.35 & 68.27 & 71.29 \\
          & DKD   & 49.50 & \textbf{69.82} & 85.26 & 91.07 & \textbf{80.00} & 71.54 & 77.84/35.68 & 73.01 & 79.09/77.65 & 79.42 & \textbf{73.20} & 70.96 & 66.35 & 71.04 \\
          & NKD   & 50.17 & 67.19 & 84.01 & \textbf{92.86} & 79.00 & 72.68 & 79.69/37.67 & 73.50 & 78.50/77.09 & 81.23 & 71.32 & \textbf{72.06} & 66.35 & 71.16 \\
          & NormKD & 48.16 & 67.54 & \textbf{85.35} & 89.29 & 79.00 & 70.57 & 77.19/35.57 & 71.82 & 78.44/76.98 & 80.87 & 72.88 & 70.48 & 68.27 & 70.52 \\
          & Top-$k$ KL & 47.49 & 68.25 & 84.19 & 87.50 & 77.00 & \textbf{72.75} & 79.39/37.67 & \textbf{74.59} & 79.40/78.01 & \textbf{82.67} & 72.10 & 70.80 & 64.42 & 70.57 \\
          & BiLD (ours) & 49.83 & 67.54 & 84.86 & 91.07 & \textbf{80.00} & 72.10 & \textbf{79.49/37.78} & 73.61 & \textbf{79.96/78.57} & \textbf{82.67} & 72.88 & 71.98 & 71.15 & \textbf{71.85} \bigstrut[b]\\
    \hline
    \multirow{8}[6]{*}{BLOOM-1B} & SFT   & 34.78 & 53.86 & 80.76 & 87.50 & 64.00 & 37.39 & 73.18/30.12 & 65.72 & 72.04/70.59 & 73.65 & 67.71 & 67.40 & 64.42 & 61.38 \bigstrut[t]\\
          & Vanilla KL & 45.48 & 64.39 & 83.67 & 87.50 & 73.00 & 65.31 & 77.66/33.37 & 70.95 & 77.11/75.67 & 77.62 & 68.03 & 68.43 & 68.27 & 67.82 \\
          & RKL   & 45.48 & \textbf{65.44} & 83.43 & 85.71 & 74.00 & 65.70 & 76.63/32.95 & 70.78 & \textbf{77.51/76.10} & 79.42 & 70.69 & 68.27 & 64.42 & 67.88 \\
          & DKD   & 42.47 & 64.56 & \textbf{84.10} & 85.71 & 72.00 & 63.72 & 75.49/31.79 & 69.48 & 75.78/74.46 & 79.78 & \textbf{71.79} & 68.98 & \textbf{69.23} & 67.55 \\
          & NKD   & 43.14 & 60.88 & 82.75 & 89.29 & 68.00 & 63.53 & 76.94/34.84 & 70.73 & 75.31/73.87 & 77.62 & 69.44 & \textbf{69.30} & 61.54 & 66.53 \\
          & NormKD & 42.81 & 61.05 & 83.82 & 83.93 & 69.00 & 62.80 & 74.13/30.75 & 67.74 & 74.49/72.95 & 77.62 & 69.91 & 67.80 & 65.38 & 65.81 \\
          & Top-$k$ KL & \textbf{49.50} & 62.11 & 83.06 & 89.29 & 74.00 & \textbf{65.72} & 78.30/34.73 & 71.22 & 77.28/75.89 & 77.98 & 70.22 & \textbf{69.30} & 60.58 & 67.97 \\
          & BiLD (ours) & 44.48 & 62.98 & 83.39 & \textbf{91.07} & \textbf{77.00} & 64.84 & \textbf{78.37/35.78} & \textbf{72.20} & 77.23/75.93 & \textbf{80.14} & 70.53 & \textbf{69.30} & 68.27 & \textbf{68.92} \bigstrut[b]\\
    \Xhline{1pt}
    Qwen-4B & Teacher & 68.23 & 81.40 & 87.43 & 96.43 & 89.00 & 86.30 & 85.85/51.63 & 82.10 & 82.59/81.10 & 87.73 & 72.73 & 80.82 & 74.04 & 79.92 \bigstrut\\
    \hline
    \multicolumn{1}{c}{\multirow{8}[6]{*}{Qwen-1.8B}} & SFT   & 52.17 & 73.86 & 83.88 & 91.07 & 86.00 & 72.58 & 79.95/39.66 & 75.90 & 77.37/76.05 & 84.12 & 71.79 & 72.06 & 61.54 & 72.36 \bigstrut[t]\\
          & Vanilla KL & \textbf{55.52} & 74.74 & 85.60 & 96.43 & 86.00 & 77.74 & 79.46/36.52 & 76.66 & 79.24/36.52 & 85.56 & 69.59 & 75.14 & 64.42 & 73.98 \\
          & RKL   & 50.84 & 76.14 & 85.14 & 94.64 & 87.00 & 77.85 & 79.52/39.14 & 76.39 & 79.49/77.98 & 84.48 & 71.47 & 76.64 & 69.23 & 74.38 \\
          & DKD   & 51.84 & \textbf{77.02} & 85.75 & \textbf{98.21} & 85.00 & 76.90 & 80.56/39.77 & 74.54 & 77.91/76.18 & 84.48 & 71.16 & 76.56 & 67.31 & 74.21 \\
          & NKD   & 51.84 & 73.33 & 84.53 & 92.86 & \textbf{88.00} & 77.49 & 81.98/42.18 & 76.61 & 79.03/77.58 & 84.12 & 70.85 & 74.98 & 66.35 & 73.90 \\
          & NormKD & 52.84 & 76.49 & 85.26 & 96.43 & 85.00 & 77.24 & 80.81/40.50 & 74.76 & 78.22/76.48 & \textbf{85.92} & 70.53 & \textbf{76.87} & \textbf{70.19} & 74.50 \\
          & Top-$k$ KL & 53.85 & 76.14 & \textbf{85.93} & 96.43 & 82.00 & \textbf{77.99} & 81.81/41.03 & 76.71 & \textbf{80.08/78.71} & 83.39 & 71.32 & 75.85 & 67.31 & 74.36 \\
          & BiLD (ours) & 54.85 & 73.16 & 84.53 & 96.43 & \textbf{88.00} & 77.56 & \textbf{81.49/42.92} & \textbf{77.97} & 79.87/78.56 & 85.56 & \textbf{72.10} & 76.01 & 68.27 & \textbf{75.07} \bigstrut[b]\\
    \hline
    \multicolumn{1}{c}{\multirow{8}[6]{*}{Qwen-0.5B}} & SFT   & 37.46 & 62.11 & 80.40 & 87.50 & 77.00 & 46.71 & 74.24/28.54 & 68.44 & 71.19/69.79 & 77.26 & 66.30 & 69.38 & 59.62 & 63.88 \bigstrut[t]\\
          & Vanilla KL & 43.14 & 63.68 & 81.74 & 85.71 & 78.00 & 66.73 & 75.97/29.07 & 71.87 & 72.55/70.91 & 79.78 & 70.53 & 71.35 & 60.58 & 67.16 \\
          & RKL   & \textbf{46.49} & 64.39 & 81.53 & 87.50 & \textbf{79.00} & 67.06 & 75.37/29.38 & 71.16 & 71.46/69.55 & \textbf{82.31} & 69.91 & 70.80 & 58.65 & 67.52 \\
          & DKD   & 40.80 & 62.98 & 82.66 & 82.14 & 77.00 & 61.03 & 72.35/26.55 & 66.87 & 65.68/63.20 & 81.59 & 70.06 & 70.64 & 61.54 & 65.16 \\
          & NKD   & 41.14 & 63.86 & 82.42 & 94.64 & 78.00 & 68.30 & 79.33/36.20 & 73.01 & 74.81/73.35 & \textbf{82.31} & 67.40 & 72.22 & 71.15 & 69.54 \\
          & NormKD & 41.14 & 61.40 & 82.72 & 83.93 & 77.00 & 62.31 & 74.13/29.07 & 68.55 & 67.17/64.79 & \textbf{82.31} & \textbf{71.16} & 71.43 & 62.50 & 66.02 \\
          & Top-$k$ KL & 43.14 & 65.79 & 82.39 & 94.64 & 77.00 & 68.58 & 78.83/35.89 & 71.82 & 74.30/72.95 & \textbf{82.31} & 69.28 & \textbf{73.24} & 62.50 & 69.19 \\
          & BiLD (ours) & 41.81 & \textbf{67.54} & \textbf{83.43} & \textbf{96.43} & 78.00 & \textbf{68.99} & \textbf{79.72/37.78} & \textbf{73.34} & \textbf{75.22/73.94} & 81.59 & 69.75 & 72.22 & \textbf{74.04} & \textbf{70.68} \bigstrut[b]\\
    \Xhline{1pt}
    \end{tabular}%
    }
    \caption{The overall performance of various distillation methods and SFT baselines, with best results shown in \textbf{bold}. When choosing the best results and calculating the Average Accuracy (Avg.), we use EM score for the MultiRC dataset and Accuracy for the ReCoRD dataset. The instruction templates for each dataset are listed in Appendix~\ref{app:templates}.}
  \label{tab:bild_results}%
\end{table*}%

\subsection{Implementation Details}
\label{sec:impl_details}

We conduct experiments using the BLOOM and Qwen1.5 (abbreviated as Qwen) models, which are available in various sizes. Specifically, we employ BLOOM-7B and Qwen-4B as teacher models. For student models, we select BLOOM-3B and BLOOM-1B from the BLOOM series, and 1.8B and 0.5B versions from Qwen.

We perform three epochs of SFT on each teacher model and eight epochs of distillation for each student. Both SFT and distillation are conducted with a batch size of 64 and a micro batch size of 2, using the full dataset. We employ a cosine scheduler with an initial learning rate of $1e-5$ for SFT and $2e-5$ for distillation. The warm-up steps are set to 64. During SFT, we utilize the cross entropy loss. For the different distillation methods we tested, all parameters, except for temperature, are set to their default values. Due to the computational complexity of some distillation methods, we use the vanilla KL loss for the instruction part to expedite the distillation process, and apply different distillation losses to the output part. The temperature $T$ for all loss functions is 3. For the top-$k$ KL loss, we set $k$=1024, and for our proposed BiLD loss, we set $k$=8. 

All our experiments are carried out on 8 NVIDIA A100 GPUs. To reduce memory usage, we employ DeepSpeed \citep{rasley2020deepspeed} during both SFT and distillation processes, along with gradient checkpointing and BFLOAT16 mode \citep{kalamkar2019study}. We have not explored the minimum memory requirements. However, in practice, experiments involving all methods except DKD \citep{zhao2022decoupled}, NKD \citep{yang2023knowledge}, and NormKD \citep{chi2023normkd} can be conducted with half of the computational resources. During the evaluation, we employ vLLM \citep{kwon2023efficient} for faster inference. The evaluation can be performed with a single NVIDIA A100 GPU. More implementation details can be found in \href{https://github.com/fpcsong/BiLD}{our repository}.

\section{Results and Analysis}
\label{sec:results_and_analysis}

\subsection{Main Results}
\label{sec:main_results}

We report the experimental results on all 13 datasets in Table~\ref{tab:bild_results}. Across all four sets of distillation, the BiLD loss achieves the highest average accuracy, outperforming SFT, vanilla KL, and the other five methods we tested. In the distillation from Qwen-4B to 0.5B, the BiLD loss showed a significant improvement in average accuracy, surpassing the vanilla KL loss by 3.52\%. This improvement is also observed in the distillation from Qwen-4B to 1.8B and from BLOOM-7B to 1B, with improvements of 1.09\% and 1.10\% over the vanilla KL loss respectively. A notable case is the distillation from BLOOM-7B to 1B, where the student using vanilla KL loss can easily match the teacher's performance. In this scenario, our BiLD loss still maintained a consistent advantage, with an average increase of 0.64\% over the vanilla KL loss. In contrast, other methods show only marginal improvements or even declines in performance. The robust performance of the BiLD loss across various distillation scenarios underscores its superiority and effectiveness.

\subsection{Analysis of the Effectiveness of Clipping Logits}
\label{sec:analysis_on_clipping}

The experimental results in Table~\ref{tab:bild_results} indicate that, in three distillation scenarios, simply clipping the full logits to the top-$k$ logits improves the performance of the KL loss. This suggests that filtering out the noise in the logits' long tail distribution can be a practical and straightforward approach to enhancing distillation performance. Our statistics show that the top-1024 logits cover over 99\% of the probability in both Qwen-4B and BLOOM-7B teachers. For computational simplicity, we set $k$=1024 for the top-$k$ KL loss to verify that excluding the long-tail distribution of logits can improve distillation results.

\begin{table}[!ht]
  \centering
  \resizebox{0.95\columnwidth}{!}{%
    \begin{tabular}{crcc}
    \Xhline{1pt}
    Model & Method & ${\rm Overlap}@1$ & ${\rm Overlap}@8$ \bigstrut\\
    \hline
    \multirow{8}[1]{*}{BLOOM-3B} & SFT   & 74.89 & 44.61 \\
          & Vanilla KL & \textbf{82.51} & 54.64 \\
          & RKL   & 82.31 & 54.64 \\
          & DKD   & 74.00 & 52.39 \\
          & NKD   & 82.11 & 53.25 \\
          & NormKD & 48.80 & 36.95 \\
          & Top-$k$ KL & 81.67 & 55.73 \\
          & BiLD  & 81.72 & \textbf{56.57} \bigstrut[b]\\
    \hline
    \multirow{8}[2]{*}{BLOOM-1B} & SFT   & 74.40 & 40.71 \bigstrut[t]\\
          & Vanilla KL & 80.82 & 51.91 \\
          & RKL   & 80.71 & 51.58 \\
          & DKD   & 75.44 & 48.83 \\
          & NKD   & 79.59 & 50.01 \\
          & NormKD & 73.56 & 42.70 \\
          & Top-$k$ KL & 80.20 & 50.87 \\
          & BiLD  & \textbf{81.21} & \textbf{52.86} \bigstrut[b]\\
    \hline
    \multicolumn{1}{c}{\multirow{8}[2]{*}{Qwen-1.8B}} & SFT   & 93.30 & 53.28 \bigstrut[t]\\
          & Vanilla KL & 94.35 & 68.02 \\
          & RKL   & 94.31 & 67.93 \\
          & DKD   & 94.09 & 67.01 \\
          & NKD   & 94.02 & 65.01 \\
          & NormKD & 94.26 & 68.32 \\
          & Top-$k$ KL & \textbf{94.43} & 67.55 \\
          & BiLD  & 94.39 & \textbf{70.97} \bigstrut[b]\\
    \hline
    \multicolumn{1}{c}{\multirow{8}[1]{*}{Qwen-0.5B}} & SFT   & 91.67 & 47.29 \bigstrut[t]\\
          & Vanilla KL & 92.72 & 61.81 \\
          & RKL   & 92.54 & 61.65 \\
          & DKD   & 91.50 & 56.62 \\
          & NKD   & 92.88 & 59.11 \\
          & NormKD & 91.76 & 58.16 \\
          & Top-$k$ KL & 93.11 & 64.00 \\
          & BiLD  & \textbf{93.23} & \textbf{68.58} \\
    \Xhline{1pt}
    \end{tabular}%
    }
  \caption{The top-1 and top-8 overlap of different distillation methods on 4 distillation settings.}
  \label{tab:overlap}%
\end{table}%

\subsection{Analysis of Performance at the Logit Level}
\label{sec:analysis_on_logits_level}

To demonstrate the performance of different distillation methods at the logit level, we introduce a new metric, top-$k$ overlap (${\rm overlap}@k$). Consider an instruction $I$ represented as a sequence of tokens. We denote the output tokens generated by the teacher with $I$ as $A^t$, and the concatenated sequence of tokens as $C^t = I \oplus A^t$. The logits sequence for the teacher's output part can be represented as $\textbf{Z}^t = \left[\textbf{z}^t_1, \textbf{z}^t_2, \cdots, \textbf{z}^t_M \right]$, where $M$ is the length of $A^t$. The element $\textbf{z}^t_i$ within $\textbf{Z}^t$ is the logits at the $i$-th position of the teacher's output part. By feeding the whole $C^t$ into the student, we denote the student logits sequence corresponding to the positions of $A^t$ as $\textbf{Z}^s = \left[\textbf{z}^s_1, \textbf{z}^s_2, \cdots, \textbf{z}^s_M \right]$. Consequently, we define the top-$k$ overlap as:

\begin{equation}
\label{eq:top_k_overlap}
{\rm overlap}@{k} = \frac{1}{M}  \sum_{i=1}^{M} \frac{{\rm topk}(\textbf{z}^t_i) \cap {\rm topk}(\textbf{z}^s_i) }{k},
\end{equation}

\noindent where ${\rm topk(\cdot)}$ is a function to select tokens corresponding to the top-$k$ logit values. The metric ${\rm overlap}@{k}$ measures the average degree of overlap for the top-$k$ logits corresponding tokens at the same positions in $\textbf{Z}^t$ and $\textbf{Z}^s$. Specifically, ${\rm overlap}@{1}$ evaluates whether the tokens with the highest logit values from the teacher and student outputs match at each position. This metric can measure the efficacy of LLMs in greedy search mode, where LLMs generate text based on the token with the highest probability. For $k > 1$, ${\rm overlap}@{k}$ calculates the ratio of overlapping tokens corresponding to the top-$k$ logits from both student and teacher at each position, reflecting how well the student imitates the important parts of teacher logits. From another perspective, ${\rm overlap}@{1}$ measures the performance of models in scenarios where there is only one correct answer, while ${\rm overlap}@{k} (k > 1)$ reflects the degree of similarity between the student and teacher responses in open-ended scenarios.

According to the results in Table~\ref{tab:overlap}, our proposed BiLD loss notably enhances ${\rm overlap}@{8}$ while maintaining a competitive ${\rm overlap}@{1}$. Compared to other methods, students trained with BiLD loss better imitate the teacher's primary behaviors at the logit level, indicating that BiLD loss helps student logits align with the important part of teacher logits.


\subsection{Ablation Study}
\label{sec:ablation_study}

\subsubsection{Impact of Temperature}
\label{sec:ablation_temperature}

\begin{figure}
\centering
    \includegraphics[width=0.8\columnwidth]{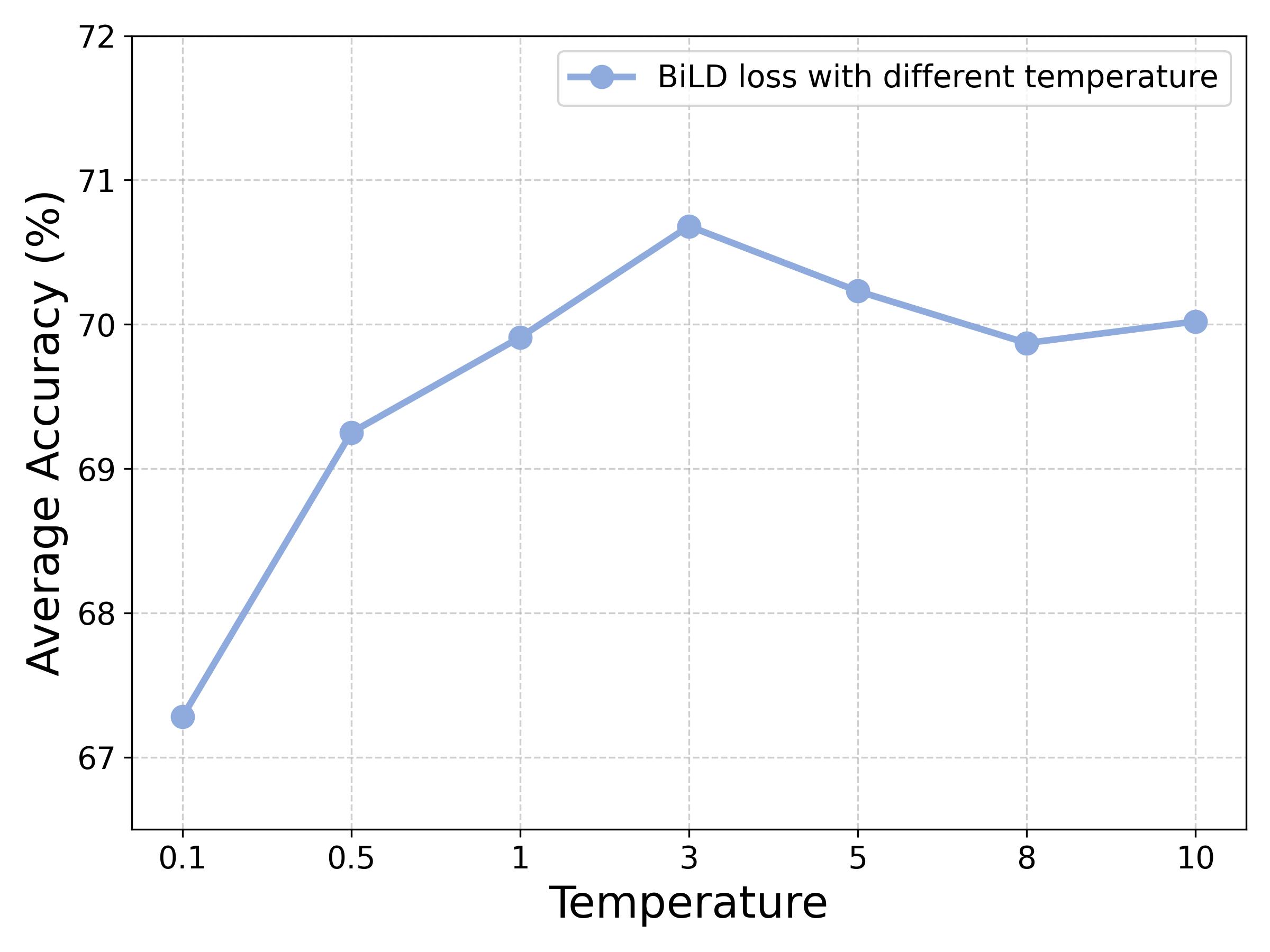}
\caption{Ablation study of model temperature.}
\label{fig:abl_temp}
\end{figure}

To understand the impact of temperature during the distillation of BiLD loss, we vary the temperature parameter $T \in \left \{ 0.1, 0.5, 1, 3, 5, 8, 10 \right \}$ while keeping other hyperparameters and model architectures constant. The experimental results, as depicted in Figure~\ref{fig:abl_temp}, indicate that lower temperatures significantly degrade the performance of BiLD loss. We choose $T$=3, which yields the best performance, for our distillation experiments.

\subsubsection{Impact of the $k$ Value in BiLD Loss}
\label{sec:ablation_bild_k}

\begin{figure}
\centering
    \includegraphics[width=0.8\columnwidth]{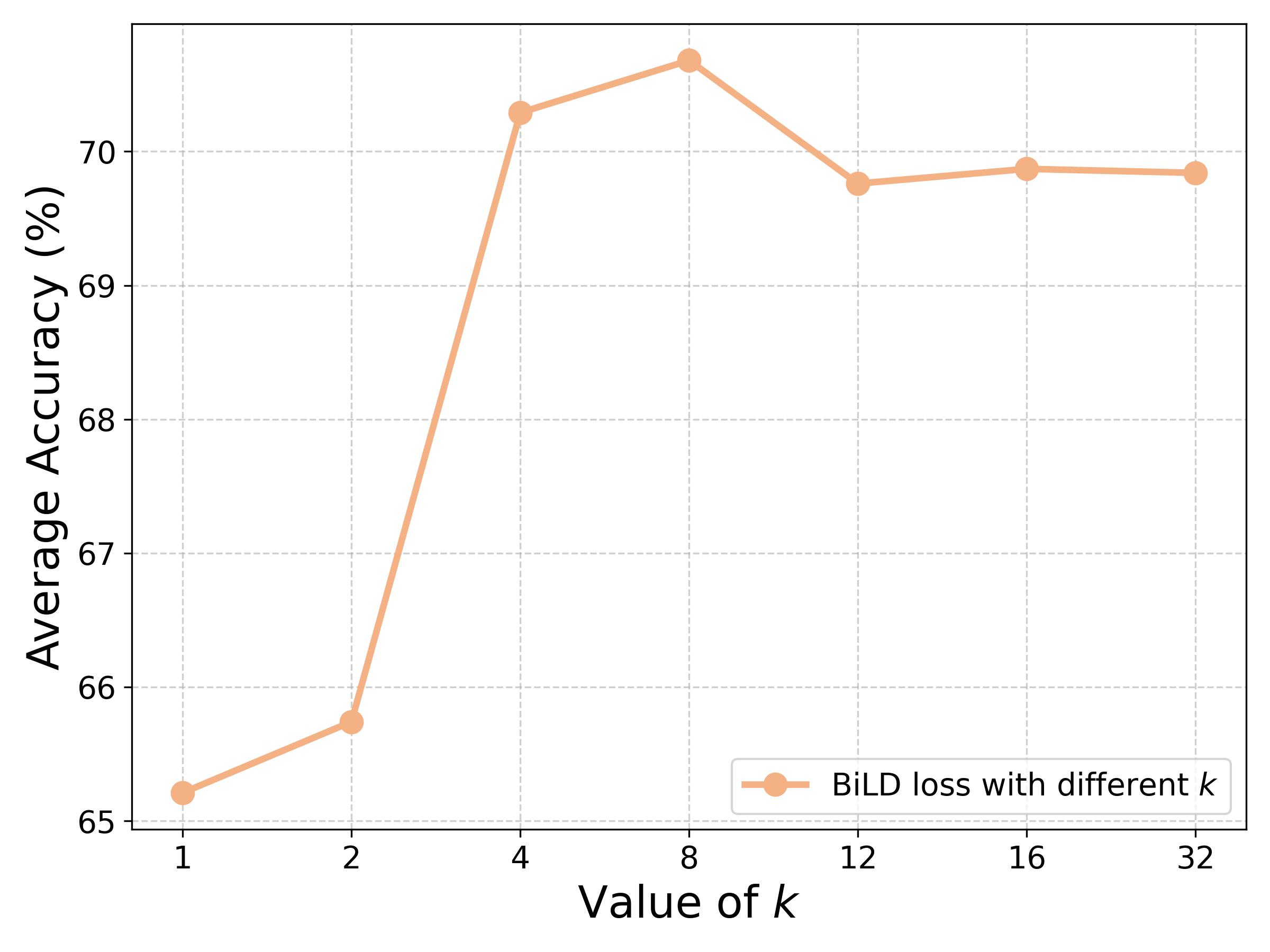}
\caption{Ablation study of $k$ values in BiLD loss.}
\label{fig:abl_bild_k}
\end{figure}

The hyperparameter $k$ controls the length of clipped logits in BiLD loss. We experiment with $k \in \left \{ 1, 2, 4, 8, 12, 16, 32 \right \} $ and evaluate the distillation results using average accuracy as well as top-1, top-8, and top-32 overlap. We report the results in Figure~\ref{fig:abl_bild_k} and Table~\ref{tab:abl_bild_k}. Smaller $k$ values ($k \in \left \{ 1, 2 \right \}$) lead to overly short logits, resulting in poor performance. As $k$ increases, both average accuracy and ${\rm overlap}@{1}$ rise and then stabilize, while significant improvements can be seen in ${\rm overlap}@{8}$ and ${\rm overlap}@{32}$. However, higher $k$ values lead to increased computational costs. Considering the trade-off between computation time and performance, we select $k$=8 for BiLD loss in our experiments.

\begin{table}[!ht]
  \centering
  \resizebox{0.95\columnwidth}{!}{%
    \begin{tabular}{cccc}
    \Xhline{1pt}
    top-$k$ & Overlap@1 & Overlap@8 & Overlap@32 \bigstrut\\
    \hline
    $k$=1   & 91.93 & 49.57 & 38.91 \bigstrut[t]\\
    $k$=2   & 91.97 & 49.60 & 38.93 \\
    $k$=4   & 93.21 & 63.64 & 47.05 \\
    $k$=8   & 93.23 & 68.58 & 52.98 \\
    $k$=12  & 93.16 & 69.46 & 56.00 \\
    $k$=16  & 93.17 & 69.56 & 57.75 \\
    $k$=32  & 93.12 & 69.29 & 60.77 \bigstrut[b]\\
    \Xhline{1pt}
    \end{tabular}%
  }
  \caption{Top-1, top-$8$ and top-$32$ overlap.}
  \label{tab:abl_bild_k}%
\end{table}%

\section{Conclusion}
\label{sec:conclusion}

In this paper, we propose the Bi-directional Logits Difference (BiLD) loss, a novel optimization objective for distilling large language models (LLMs). The BiLD loss enhances distillation performance by filtering out long-tail noise and leveraging internal logits ranking information. It achieves superior distillation performance using only the top-8 logits compared to vanilla KL loss using full logits and other distillation methods. Our extensive experiments across diverse datasets and model architectures confirm the effectiveness of the BiLD loss, demonstrating its ability to more efficiently capture key knowledge from the teacher model.

\section*{Limitations}
\label{sec:limitations}

Our approach falls within the realm of logits distillation, necessitating access to teacher logits. However, powerful LLMs such as GPT-4 \citep{achiam2023gpt} and Gemini \citep{team2023gemini} currently provide only output text or incomplete logits access, making our method unable to utilize these highly capable LLMs as teachers. Additionally, our Bi-directional Logits Difference (BiLD) loss requires shared vocabularies between the teacher and student models to ensure vector space alignment.

Another challenge lies in the computational complexity of our BiLD loss, particularly during the construction of logits differences using top-$k$ logits. Although we demonstrate that using only the top-8 logits achieves better results than the vanilla KL loss, increasing the number of clipped logits leads to a rapid escalation in our method's time overhead, which becomes a practical concern.

Furthermore, our approach directly clips the long-tail part of logits during distillation. While this approach improves performance, it unavoidably results in the loss of knowledge contained within the long-tail distribution. Investigating methods to better utilize the knowledge hidden in the long-tail distribution represents a promising avenue for future research.



\bibliography{custom}

\begin{thebibliography}{44}
\providecommand{\natexlab}[1]{#1}

\bibitem[{Achiam et~al.(2023)Achiam, Adler, Agarwal, Ahmad, Akkaya, Aleman, Almeida, Altenschmidt, Altman, Anadkat et~al.}]{achiam2023gpt}
Josh Achiam, Steven Adler, Sandhini Agarwal, Lama Ahmad, Ilge Akkaya, Florencia~Leoni Aleman, Diogo Almeida, Janko Altenschmidt, Sam Altman, Shyamal Anadkat, et~al. 2023.
\newblock Gpt-4 technical report.
\newblock \emph{arXiv preprint arXiv:2303.08774}.

\bibitem[{Agarwal et~al.(2024)Agarwal, Vieillard, Zhou, Stanczyk, Garea, Geist, and Bachem}]{agarwal2024policy}
Rishabh Agarwal, Nino Vieillard, Yongchao Zhou, Piotr Stanczyk, Sabela~Ramos Garea, Matthieu Geist, and Olivier Bachem. 2024.
\newblock On-policy distillation of language models: Learning from self-generated mistakes.
\newblock In \emph{The Twelfth International Conference on Learning Representations}.

\bibitem[{Bai et~al.(2023)Bai, Bai, Chu, Cui, Dang, Deng, Fan, Ge, Han, Huang, Hui, Ji, Li, Lin, Lin, Liu, Liu, Lu, Lu, Ma, Men, Ren, Ren, Tan, Tan, Tu, Wang, Wang, Wang, Wu, Xu, Xu, Yang, Yang, Yang, Yang, Yao, Yu, Yuan, Yuan, Zhang, Zhang, Zhang, Zhang, Zhou, Zhou, Zhou, and Zhu}]{qwen}
Jinze Bai, Shuai Bai, Yunfei Chu, Zeyu Cui, Kai Dang, Xiaodong Deng, Yang Fan, Wenbin Ge, Yu~Han, Fei Huang, Binyuan Hui, Luo Ji, Mei Li, Junyang Lin, Runji Lin, Dayiheng Liu, Gao Liu, Chengqiang Lu, Keming Lu, Jianxin Ma, Rui Men, Xingzhang Ren, Xuancheng Ren, Chuanqi Tan, Sinan Tan, Jianhong Tu, Peng Wang, Shijie Wang, Wei Wang, Shengguang Wu, Benfeng Xu, Jin Xu, An~Yang, Hao Yang, Jian Yang, Shusheng Yang, Yang Yao, Bowen Yu, Hongyi Yuan, Zheng Yuan, Jianwei Zhang, Xingxuan Zhang, Yichang Zhang, Zhenru Zhang, Chang Zhou, Jingren Zhou, Xiaohuan Zhou, and Tianhang Zhu. 2023.
\newblock Qwen technical report.
\newblock \emph{arXiv preprint arXiv:2309.16609}.

\bibitem[{{BigScience Workshop}(2022)}]{bigscience_workshop_2022}
{BigScience Workshop}. 2022.
\newblock \href {https://doi.org/10.57967/hf/0003} {Bloom (revision 4ab0472)}.

\bibitem[{Bisk et~al.(2020)Bisk, Zellers, Gao, Choi et~al.}]{bisk2020piqa}
Yonatan Bisk, Rowan Zellers, Jianfeng Gao, Yejin Choi, et~al. 2020.
\newblock Piqa: Reasoning about physical commonsense in natural language.
\newblock In \emph{Proceedings of the AAAI conference on artificial intelligence}, volume~34, pages 7432--7439.

\bibitem[{Chan et~al.(2022)Chan, Silva, Lim, Kozuno, Mahmood, and White}]{chan2022greedification}
Alan Chan, Hugo Silva, Sungsu Lim, Tadashi Kozuno, A~Rupam Mahmood, and Martha White. 2022.
\newblock Greedification operators for policy optimization: Investigating forward and reverse kl divergences.
\newblock \emph{Journal of Machine Learning Research}, 23(253):1--79.

\bibitem[{Chi et~al.(2023)Chi, Zheng, Li, Yang, Wu, Lin, and Cai}]{chi2023normkd}
Zhihao Chi, Tu~Zheng, Hengjia Li, Zheng Yang, Boxi Wu, Binbin Lin, and Deng Cai. 2023.
\newblock Normkd: Normalized logits for knowledge distillation.
\newblock \emph{arXiv preprint arXiv:2308.00520}.

\bibitem[{Clark et~al.(2019)Clark, Lee, Chang, Kwiatkowski, Collins, and Toutanova}]{clark2019boolq}
Christopher Clark, Kenton Lee, Ming-Wei Chang, Tom Kwiatkowski, Michael Collins, and Kristina Toutanova. 2019.
\newblock Boolq: Exploring the surprising difficulty of natural yes/no questions.
\newblock \emph{arXiv preprint arXiv:1905.10044}.

\bibitem[{Clark et~al.(2018)Clark, Cowhey, Etzioni, Khot, Sabharwal, Schoenick, and Tafjord}]{clark2018think}
Peter Clark, Isaac Cowhey, Oren Etzioni, Tushar Khot, Ashish Sabharwal, Carissa Schoenick, and Oyvind Tafjord. 2018.
\newblock Think you have solved question answering? try arc, the ai2 reasoning challenge.
\newblock \emph{arXiv preprint arXiv:1803.05457}.

\bibitem[{Cui et~al.(2024)Cui, Qin, Gao, Zhang, Xu, Wu, Li, Sun, Zhou, and Li}]{cui2024sinkhorn}
Xiao Cui, Yulei Qin, Yuting Gao, Enwei Zhang, Zihan Xu, Tong Wu, Ke~Li, Xing Sun, Wengang Zhou, and Houqiang Li. 2024.
\newblock Sinkhorn distance minimization for knowledge distillation.
\newblock \emph{arXiv preprint arXiv:2402.17110}.

\bibitem[{De~Marneffe et~al.(2019)De~Marneffe, Simons, and Tonhauser}]{de2019commitmentbank}
Marie-Catherine De~Marneffe, Mandy Simons, and Judith Tonhauser. 2019.
\newblock The commitmentbank: Investigating projection in naturally occurring discourse.
\newblock In \emph{proceedings of Sinn und Bedeutung}, volume~23, pages 107--124.

\bibitem[{Egiazarian et~al.(2024)Egiazarian, Panferov, Kuznedelev, Frantar, Babenko, and Alistarh}]{egiazarian2024extreme}
Vage Egiazarian, Andrei Panferov, Denis Kuznedelev, Elias Frantar, Artem Babenko, and Dan Alistarh. 2024.
\newblock Extreme compression of large language models via additive quantization.
\newblock \emph{arXiv preprint arXiv:2401.06118}.

\bibitem[{Fu et~al.(2023)Fu, Peng, Ou, Sabharwal, and Khot}]{fu2023specializing}
Yao Fu, Hao Peng, Litu Ou, Ashish Sabharwal, and Tushar Khot. 2023.
\newblock Specializing smaller language models towards multi-step reasoning.
\newblock In \emph{International Conference on Machine Learning}, pages 10421--10430. PMLR.

\bibitem[{Giampiccolo et~al.(2007)Giampiccolo, Magnini, Dagan, and Dolan}]{giampiccolo2007third}
Danilo Giampiccolo, Bernardo Magnini, Ido Dagan, and William~B Dolan. 2007.
\newblock The third pascal recognizing textual entailment challenge.
\newblock In \emph{Proceedings of the ACL-PASCAL workshop on textual entailment and paraphrasing}, pages 1--9.

\bibitem[{Gu et~al.(2023)Gu, Dong, Wei, and Huang}]{gu2023minillm}
Yuxian Gu, Li~Dong, Furu Wei, and Minlie Huang. 2023.
\newblock Minillm: Knowledge distillation of large language models.
\newblock In \emph{The Twelfth International Conference on Learning Representations}.

\bibitem[{He et~al.(2016)He, Zhang, Ren, and Sun}]{he2016deep}
Kaiming He, Xiangyu Zhang, Shaoqing Ren, and Jian Sun. 2016.
\newblock Deep residual learning for image recognition.
\newblock In \emph{Proceedings of the IEEE conference on computer vision and pattern recognition}, pages 770--778.

\bibitem[{Hinton et~al.(2015)Hinton, Vinyals, and Dean}]{hinton2015distilling}
Geoffrey Hinton, Oriol Vinyals, and Jeff Dean. 2015.
\newblock Distilling the knowledge in a neural network.
\newblock \emph{arXiv preprint arXiv:1503.02531}.

\bibitem[{Huang et~al.(2022)Huang, You, Wang, Qian, and Xu}]{huang2022knowledge}
Tao Huang, Shan You, Fei Wang, Chen Qian, and Chang Xu. 2022.
\newblock Knowledge distillation from a stronger teacher.
\newblock \emph{Advances in Neural Information Processing Systems}, 35:33716--33727.

\bibitem[{Jin et~al.(2023)Jin, Wang, and Lin}]{jin2023multi}
Ying Jin, Jiaqi Wang, and Dahua Lin. 2023.
\newblock Multi-level logit distillation.
\newblock In \emph{Proceedings of the IEEE/CVF Conference on Computer Vision and Pattern Recognition}, pages 24276--24285.

\bibitem[{Kalamkar et~al.(2019)Kalamkar, Mudigere, Mellempudi, Das, Banerjee, Avancha, Vooturi, Jammalamadaka, Huang, Yuen et~al.}]{kalamkar2019study}
Dhiraj Kalamkar, Dheevatsa Mudigere, Naveen Mellempudi, Dipankar Das, Kunal Banerjee, Sasikanth Avancha, Dharma~Teja Vooturi, Nataraj Jammalamadaka, Jianyu Huang, Hector Yuen, et~al. 2019.
\newblock A study of bfloat16 for deep learning training.
\newblock \emph{arXiv preprint arXiv:1905.12322}.

\bibitem[{Khashabi et~al.(2018)Khashabi, Chaturvedi, Roth, Upadhyay, and Roth}]{khashabi2018looking}
Daniel Khashabi, Snigdha Chaturvedi, Michael Roth, Shyam Upadhyay, and Dan Roth. 2018.
\newblock Looking beyond the surface: A challenge set for reading comprehension over multiple sentences.
\newblock In \emph{Proceedings of the 2018 Conference of the North American Chapter of the Association for Computational Linguistics: Human Language Technologies, Volume 1 (Long Papers)}, pages 252--262.

\bibitem[{Ko et~al.(2024)Ko, Kim, Chen, and Yun}]{ko2024distillm}
Jongwoo Ko, Sungnyun Kim, Tianyi Chen, and Se-Young Yun. 2024.
\newblock Distillm: Towards streamlined distillation for large language models.
\newblock \emph{arXiv preprint arXiv:2402.03898}.

\bibitem[{Kwon et~al.(2023)Kwon, Li, Zhuang, Sheng, Zheng, Yu, Gonzalez, Zhang, and Stoica}]{kwon2023efficient}
Woosuk Kwon, Zhuohan Li, Siyuan Zhuang, Ying Sheng, Lianmin Zheng, Cody~Hao Yu, Joseph~E. Gonzalez, Hao Zhang, and Ion Stoica. 2023.
\newblock Efficient memory management for large language model serving with pagedattention.
\newblock In \emph{Proceedings of the ACM SIGOPS 29th Symposium on Operating Systems Principles}.

\bibitem[{Lee et~al.(2023)Lee, Park, Seo, and Kang}]{lee2023self}
Hyoje Lee, Yeachan Park, Hyun Seo, and Myungjoo Kang. 2023.
\newblock Self-knowledge distillation via dropout.
\newblock \emph{Computer Vision and Image Understanding}, 233:103720.

\bibitem[{Levesque et~al.(2012)Levesque, Davis, and Morgenstern}]{levesque2012winograd}
Hector Levesque, Ernest Davis, and Leora Morgenstern. 2012.
\newblock The winograd schema challenge.
\newblock In \emph{Thirteenth international conference on the principles of knowledge representation and reasoning}.

\bibitem[{Li and Farnia(2023)}]{li2023mode}
Cheuk~Ting Li and Farzan Farnia. 2023.
\newblock Mode-seeking divergences: theory and applications to gans.
\newblock In \emph{International Conference on Artificial Intelligence and Statistics}, pages 8321--8350. PMLR.

\bibitem[{Li et~al.(2023)Li, Hessel, Yu, Ren, Chang, and Choi}]{li2023symbolic}
Liunian~Harold Li, Jack Hessel, Youngjae Yu, Xiang Ren, Kai-Wei Chang, and Yejin Choi. 2023.
\newblock Symbolic chain-of-thought distillation: Small models can also" think" step-by-step.
\newblock \emph{arXiv preprint arXiv:2306.14050}.

\bibitem[{Liang et~al.(2023)Liang, Zuo, Zhang, He, Chen, and Zhao}]{liang2023less}
Chen Liang, Simiao Zuo, Qingru Zhang, Pengcheng He, Weizhu Chen, and Tuo Zhao. 2023.
\newblock Less is more: Task-aware layer-wise distillation for language model compression.
\newblock In \emph{International Conference on Machine Learning}, pages 20852--20867. PMLR.

\bibitem[{Ma et~al.(2023)Ma, Fang, and Wang}]{ma2023llm}
Xinyin Ma, Gongfan Fang, and Xinchao Wang. 2023.
\newblock Llm-pruner: On the structural pruning of large language models.
\newblock \emph{Advances in neural information processing systems}, 36:21702--21720.

\bibitem[{Pilehvar and Camacho-Collados(2018)}]{pilehvar2018wic}
Mohammad~Taher Pilehvar and Jose Camacho-Collados. 2018.
\newblock Wic: the word-in-context dataset for evaluating context-sensitive meaning representations.
\newblock \emph{arXiv preprint arXiv:1808.09121}.

\bibitem[{Rasley et~al.(2020)Rasley, Rajbhandari, Ruwase, and He}]{rasley2020deepspeed}
Jeff Rasley, Samyam Rajbhandari, Olatunji Ruwase, and Yuxiong He. 2020.
\newblock Deepspeed: System optimizations enable training deep learning models with over 100 billion parameters.
\newblock In \emph{Proceedings of the 26th ACM SIGKDD International Conference on Knowledge Discovery \& Data Mining}, pages 3505--3506.

\bibitem[{Roemmele et~al.(2011)Roemmele, Bejan, and Gordon}]{roemmele2011choice}
Melissa Roemmele, Cosmin~Adrian Bejan, and Andrew~S Gordon. 2011.
\newblock Choice of plausible alternatives: An evaluation of commonsense causal reasoning.
\newblock In \emph{2011 AAAI Spring Symposium Series}.

\bibitem[{Sahu et~al.(2023)Sahu, Vechtomova, Bahdanau, and Laradji}]{sahu2023promptmix}
Gaurav Sahu, Olga Vechtomova, Dzmitry Bahdanau, and Issam~H Laradji. 2023.
\newblock Promptmix: A class boundary augmentation method for large language model distillation.
\newblock \emph{arXiv preprint arXiv:2310.14192}.

\bibitem[{Sakaguchi et~al.(2021)Sakaguchi, Bras, Bhagavatula, and Choi}]{sakaguchi2021winogrande}
Keisuke Sakaguchi, Ronan~Le Bras, Chandra Bhagavatula, and Yejin Choi. 2021.
\newblock Winogrande: An adversarial winograd schema challenge at scale.
\newblock \emph{Communications of the ACM}, 64(9):99--106.

\bibitem[{Sun et~al.(2024)Sun, Ren, Li, Wang, and Cao}]{Sun2024Logit}
Shangquan Sun, Wenqi Ren, Jingzhi Li, Rui Wang, and Xiaochun Cao. 2024.
\newblock Logit standardization in knowledge distillation.
\newblock In \emph{Proceedings of the IEEE/CVF Conference on Computer Vision and Pattern Recognition (CVPR)}.

\bibitem[{Team et~al.(2023)Team, Anil, Borgeaud, Wu, Alayrac, Yu, Soricut, Schalkwyk, Dai, Hauth et~al.}]{team2023gemini}
Gemini Team, Rohan Anil, Sebastian Borgeaud, Yonghui Wu, Jean-Baptiste Alayrac, Jiahui Yu, Radu Soricut, Johan Schalkwyk, Andrew~M Dai, Anja Hauth, et~al. 2023.
\newblock Gemini: a family of highly capable multimodal models.
\newblock \emph{arXiv preprint arXiv:2312.11805}.

\bibitem[{Tu et~al.(2020)Tu, Pang, Wiseman, and Gimpel}]{tu2020engine}
Lifu Tu, Richard~Yuanzhe Pang, Sam Wiseman, and Kevin Gimpel. 2020.
\newblock Engine: Energy-based inference networks for non-autoregressive machine translation.
\newblock \emph{arXiv preprint arXiv:2005.00850}.

\bibitem[{Wang et~al.(2019)Wang, Pruksachatkun, Nangia, Singh, Michael, Hill, Levy, and Bowman}]{wang2019superglue}
Alex Wang, Yada Pruksachatkun, Nikita Nangia, Amanpreet Singh, Julian Michael, Felix Hill, Omer Levy, and Samuel Bowman. 2019.
\newblock Superglue: A stickier benchmark for general-purpose language understanding systems.
\newblock \emph{Advances in neural information processing systems}, 32.

\bibitem[{Yang et~al.(2023{\natexlab{a}})Yang, Xie, Zong, Feng, Niu, Sugiyama, and Huang}]{yang2023multi}
Penghui Yang, Ming-Kun Xie, Chen-Chen Zong, Lei Feng, Gang Niu, Masashi Sugiyama, and Sheng-Jun Huang. 2023{\natexlab{a}}.
\newblock Multi-label knowledge distillation.
\newblock In \emph{Proceedings of the IEEE/CVF International Conference on Computer Vision}, pages 17271--17280.

\bibitem[{Yang et~al.(2023{\natexlab{b}})Yang, Zeng, Li, Zhang, Yuan, and Li}]{yang2023knowledge}
Zhendong Yang, Ailing Zeng, Zhe Li, Tianke Zhang, Chun Yuan, and Yu~Li. 2023{\natexlab{b}}.
\newblock From knowledge distillation to self-knowledge distillation: A unified approach with normalized loss and customized soft labels.
\newblock In \emph{Proceedings of the IEEE/CVF International Conference on Computer Vision}, pages 17185--17194.

\bibitem[{Yuan et~al.(2024)Yuan, Shang, Zhou, Dong, Xue, Wu, Li, Gu, Lee, Yan et~al.}]{yuan2024llm}
Zhihang Yuan, Yuzhang Shang, Yang Zhou, Zhen Dong, Chenhao Xue, Bingzhe Wu, Zhikai Li, Qingyi Gu, Yong~Jae Lee, Yan Yan, et~al. 2024.
\newblock Llm inference unveiled: Survey and roofline model insights.
\newblock \emph{arXiv preprint arXiv:2402.16363}.

\bibitem[{Zellers et~al.(2019)Zellers, Holtzman, Bisk, Farhadi, and Choi}]{zellers2019hellaswag}
Rowan Zellers, Ari Holtzman, Yonatan Bisk, Ali Farhadi, and Yejin Choi. 2019.
\newblock Hellaswag: Can a machine really finish your sentence?
\newblock \emph{arXiv preprint arXiv:1905.07830}.

\bibitem[{Zhang et~al.(2018)Zhang, Liu, Liu, Gao, Duh, and Van~Durme}]{zhang2018record}
Sheng Zhang, Xiaodong Liu, Jingjing Liu, Jianfeng Gao, Kevin Duh, and Benjamin Van~Durme. 2018.
\newblock Record: Bridging the gap between human and machine commonsense reading comprehension.
\newblock \emph{arXiv preprint arXiv:1810.12885}.

\bibitem[{Zhao et~al.(2022)Zhao, Cui, Song, Qiu, and Liang}]{zhao2022decoupled}
Borui Zhao, Quan Cui, Renjie Song, Yiyu Qiu, and Jiajun Liang. 2022.
\newblock Decoupled knowledge distillation.
\newblock In \emph{Proceedings of the IEEE/CVF Conference on computer vision and pattern recognition}, pages 11953--11962.

\end{thebibliography}

\appendix

\section{Calculation Efficiency of BiLD}
\label{app:calc_efficiency}

\begin{figure}[!ht]
\centering
    \includegraphics[width=0.95\columnwidth]{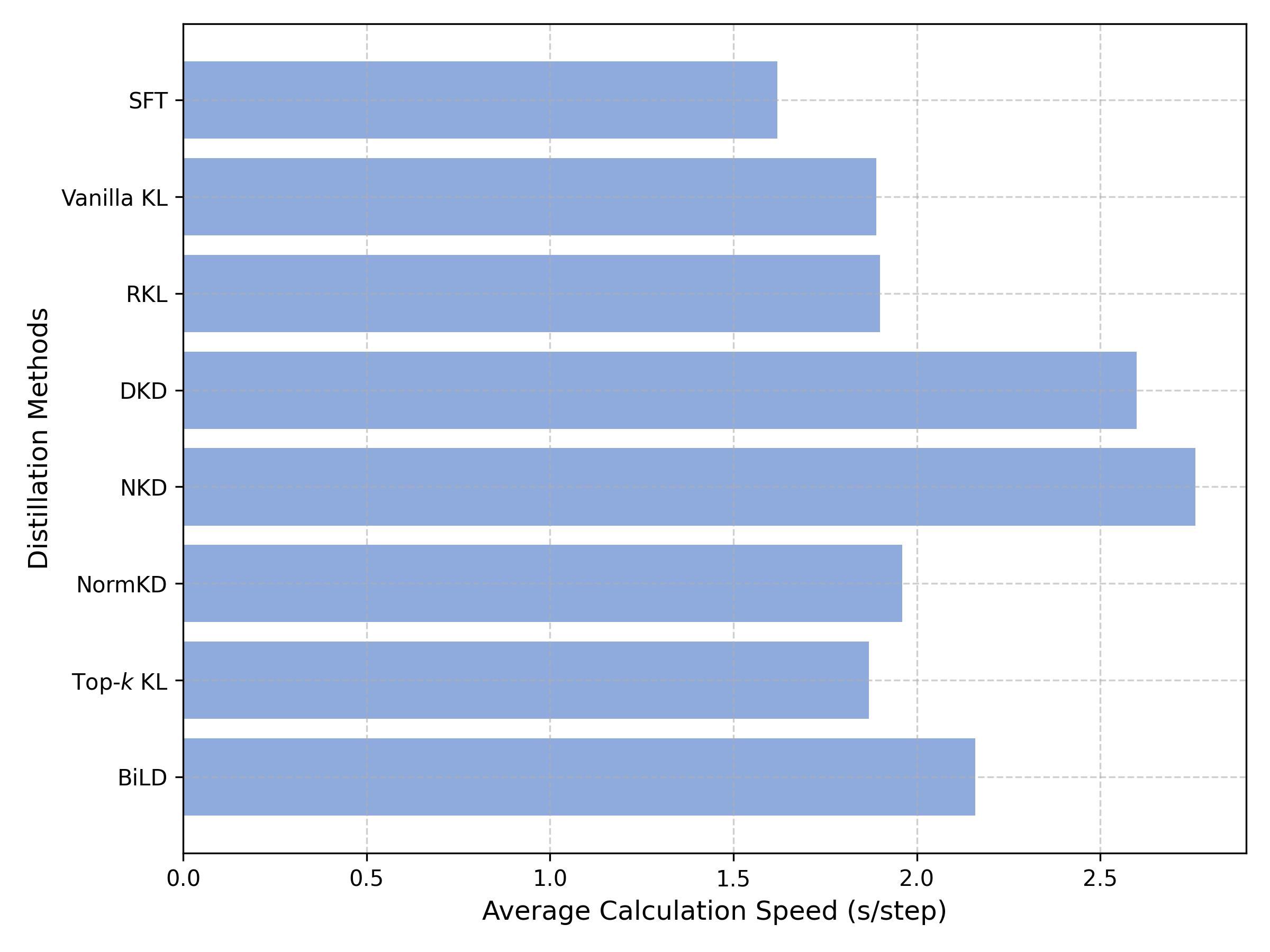}
\caption{The average calculation speed of different distillation methods.}
\label{fig:calc_time_analysis}
\end{figure}

In Figure~\ref{fig:calc_time_analysis}, we visualize the distillation speed of various methods during the distillation from Qwen-4B to 0.5B. Compared to the vanilla KL loss, our BiLD loss achieves better distillation performance with an acceptable increase in training time. Among all methods, DKD \citep{zhao2022decoupled} and NKD \citep{yang2023knowledge}, which are designed for vision models, have the slowest computation speeds due to the calculation of numerous intermediate variables. In contrast, the computation speeds of RKL, NormKD, and top-$k$ KL are comparable to the vanilla KL loss.

In the code implementation, the BiLD loss consists of two main steps: selecting the top-$k$ logit values and calculating the internal pairwise differences. Our analysis reveals that the latter step is where the significant time expenditure occurs. The time complexity for computing the internal pairwise differences is $\mathcal{O}(n^2)$, and it frequently necessitates extracting values from the tensor. This has become the time bottleneck for the BiLD loss.

\section{Details about Dataset Sizes}
\label{app:details_about_datasets}

\begin{figure}[!ht]
\centering
    \includegraphics[width=\columnwidth]{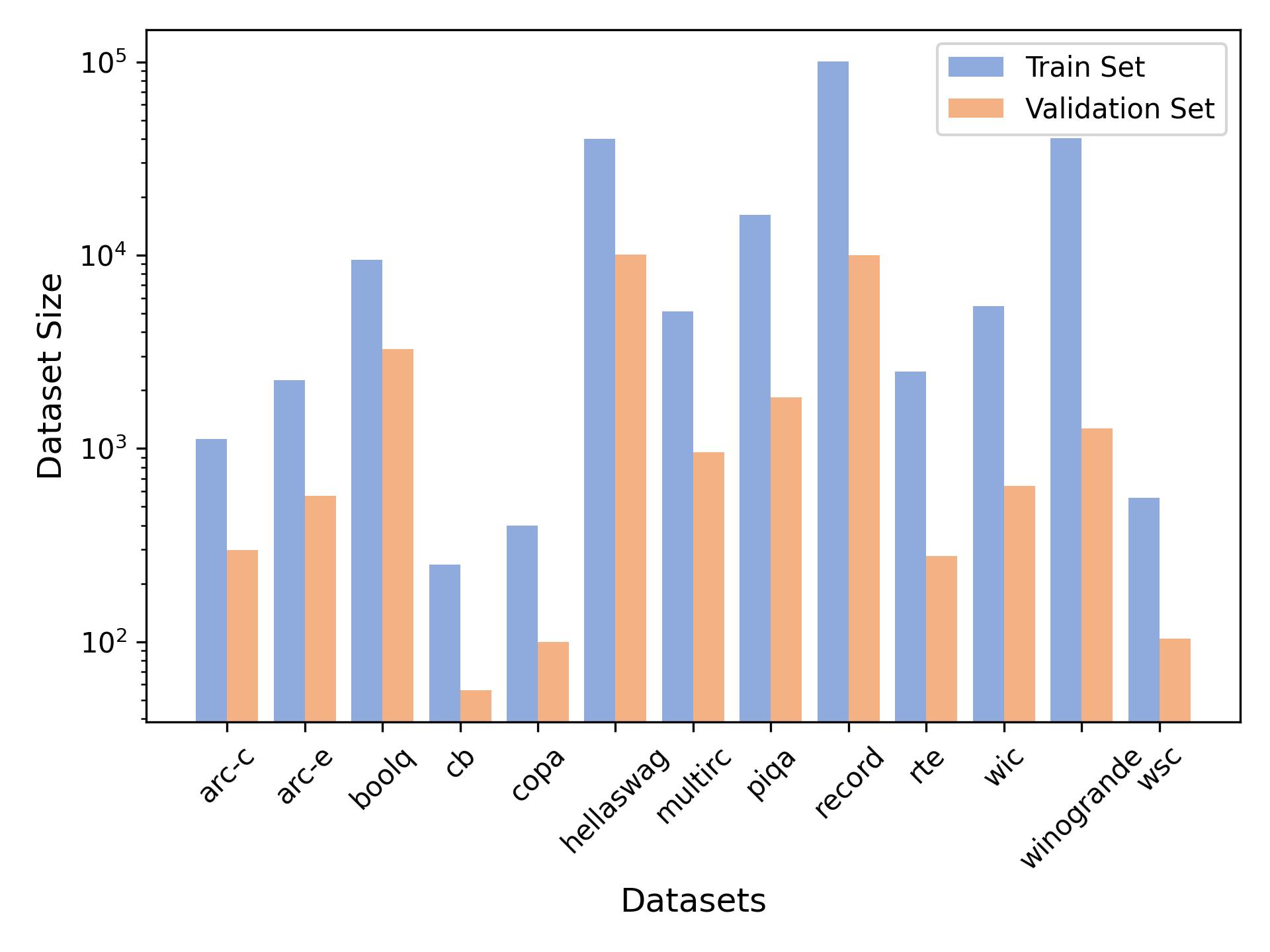}
\caption{A visualization of the dataset sizes.}
\label{fig:dataset_analysis}
\end{figure}

Details about the dataset sizes are shown in Figure \ref{fig:dataset_analysis}. There are significant size differences among the datasets, with the smallest datasets (CB, COPA, WSC) differing by three orders of magnitude from the largest dataset (ReCoRD).

\section{Toy Experiment to Compare Vision Model and LLMs' Logits}
\label{app:toy_case}

\begin{figure}[htbp]
	\centering
	\subfigure[cat.jpg] {\includegraphics[width=.18\textwidth]{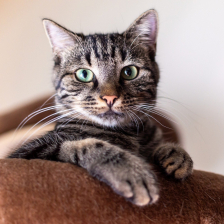}}\hspace{1em}%
	\subfigure[dogs.jpg] {\includegraphics[width=.18\textwidth]{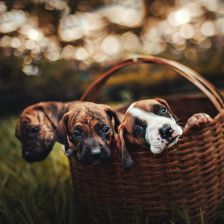}}\hspace{1em}%
	\subfigure[lioness.jpg] {\includegraphics[width=.18\textwidth]{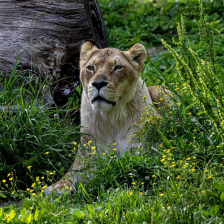}}\hspace{1em}%
	\subfigure[mushroom.jpg] {\includegraphics[width=.18\textwidth]{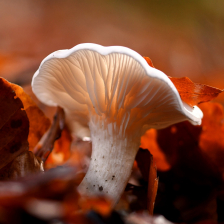}}\hspace{1em}%
	\subfigure[hat.jpg] {\includegraphics[width=.18\textwidth]{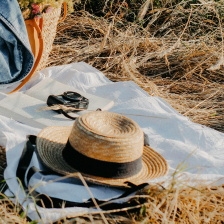}}\hspace{1em}%
	\caption{Five images used in the toy experiment. All these images are under the CC0 license, meaning they can be used for scientific purposes without risk.}
	\label{fig:toy_images}
\end{figure}

The five images we used in the toy experiments are shown in Figure~\ref{fig:toy_images}, and the five sets of instructions are in Table~\ref{tab:toy_prompts}.

\onecolumn

\begin{longtable}[htbp]{|>{\centering\arraybackslash} m{.14\textwidth} | m{.8\textwidth} |} 
\hline
& \multicolumn{1}{c|}{Instructions Content} \bigstrut\\
\hline
\label{app:ins_1}
Instruction 1 & Question: A mass of air is at an elevation of 1000 meters in the low pressure center of a Northern Hemisphere storm. Which of the following best describes the motion of air particles in this air mass due to storm conditions and the rotation of Earth as the air mass moves outward?\newline{}Choices: ['Air particles move up and to the left.', 'Air particles move up and to the right.', 'Air particles move down and to the left.', 'Air particles move down and to the right.']\newline{}Answer: \bigstrut\\
\hline
\label{app:ins_2}
Instruction 2 & Premise: A: No, I don't either. B: Uh, I mean it's, you know it, A: I don't think it's going to change very much\newline{}Hypothesis: it's going to change very much\newline{}Question: Determine whether the premise entails the hypothesis or not.\newline{}Choices: ['entailment', 'neutral', 'contradiction']\newline{}Answer: \bigstrut\\
\hline
\label{app:ins_3}
Instruction 3 & Goal: Keep laptop from overheating.\newline{}Choose the most sensible solution to achieve the goal. Choices: ['Use on top of egg carton.', 'Use on top of egg shells.']\newline{}Answer: \bigstrut\\
\hline
\label{app:ins_4}
Instruction 4 & Choose the most sensible text to replace the "\_" in the following sentence: Kyle asked Brett for some tips on healthy eating because \_ has recently lost weight.\newline{}Choices: ['Kyle', 'Brett']\newline{}Answer: \bigstrut\\
\hline
\label{app:ins_5}
Instruction 5 & Meanwhile, in the forest, the elephants are calling and hunting high and low for Arthur and Celeste , and their mothers are very worried. Fortunately, in flying over the town, an old marabou bird has seen them and come back quickly to tell the news.\newline{}Question: In the above text, does 'their' refer to 'their mothers'?\newline{}Choices:['true', 'false']\newline{}Answer: \bigstrut\\
\hline
\caption{Five instructions used in the toy experiment.}
\label{tab:toy_prompts}
\end{longtable}
\twocolumn

\onecolumn
\section{Templates}
\label{app:templates}

The template of each dataset can be seen in Table~\ref{tab:templates}.

\begin{longtable}{|>{\centering\arraybackslash} m{.14\textwidth} | m{.8\textwidth} |} 
\hline
Dataset & \multicolumn{1}{c|}{Template} \bigstrut\\
\hline
Arc-C & Question: A scientist is measuring the amount of movement along a fault. Which tool is best used for making this measurement?\newline{}Choices: ['barometer', 'stopwatch', 'meter stick', 'magnifying lens']\newline{}Answer: \bigstrut\\
\hline
Arc-E & Question: Which color shirt will reflect the most light on a hot, sunny day?\newline{}Choices: ['black', 'blue', 'red', 'white']\newline{}Answer: \bigstrut\\
\hline
BoolQ & Turn on red – Right turns on red are permitted in many regions of North America. While Western states have allowed it for more than 50 years; eastern states amended their traffic laws to allow it in the 1970s as a fuel-saving measure in response to motor fuel shortages in 1973. The Energy Policy and Conservation Act of 1975 required in §362(c)(5) that in order for a state to receive federal assistance in developing mandated conservation programs, they must permit right turns on red lights. All 50 states, the District of Columbia, Guam, and Puerto Rico have allowed right turns on red since 1980, except where prohibited by a sign or where right turns are controlled by dedicated traffic lights. (On January 1, 1980, Massachusetts became the last US state to allow right turns on red.) The few exceptions include New York City, where right turns on red are prohibited, unless a sign indicates otherwise.\newline{}Question: is it legal to turn right on red in california?\newline{}Choices: ['true', 'false']\newline{}Answer: \bigstrut\\
\hline
CB    & Premise: B: And I've worked in the hospital for fifteen years and I've taken care of a few AIDS patients. A: Uh-huh. B: Uh, when they asked us did we want to, uh, keep it the same or, uh, spend more, spend less, uh, I think right now what they're spending is adequate. Uh, for my personal opinion. Uh, because I think it's something that's going to take them a while to come up with a, uh, vaccine for. A: Yeah. Uh-huh. Uh-huh. B: I don't think it's going to be that easy to come up with\newline{}Hypothesis: it is going to be that easy to come up with\newline{}Question: Determine whether the premise entails the hypothesis or not.\newline{}Choices: ['entailment', 'neutral', 'contradiction']\newline{}Answer: \bigstrut\\
\hline
COPA  & Premise: The woman betrayed her friend.\newline{}Question: What could be the possible effect of the premise?\newline{}Choices: ['Her friend sent her a greeting card.', 'Her friend cut off contact with her.']\newline{}Your answer: \bigstrut\\
\hline
HellaSwag & Please choose the most appropriate text to complete the passage below:\newline{}Passage: [header] How to clean a plastic retainer [title] Rinse the retainer with warm or cold water. [step] The water will prep your retainer for the cleaning process. [title] Apply a mild soap onto a toothbrush.\newline{}Choices: ['[step] Rinse the retainer under the faucet bowl with warm water. Suds will accumulate on the toothbrush.', '[step] Rinse the retainer slowly from top to bottom and then wipe it on the toothbrush. Soap can effectively clean a plastic retainer but it can potentially cause irritation.', '[step] If you are using an old toothbrush, you may brush the bristles for pleasure. Fill a bucket, then fill it with a cup of liquid soap.', '[step] You can use liquid castile soap or a mild dishwashing detergent. Additionally, use a soft-bristled toothbrush.']\newline{}Answer: \bigstrut\\
\hline
MultiRC & Passage: One of the most dramatic changes in priorities proposed by the City Council would shift \$25.6 million from funding for court-appointed lawyers to the Legal Aid Society. In a document released yesterday to justify its reordered priorities, the Council contended that Legal Aid can achieve greater economies of scale than lawyers appointed pursuant to Article 18-B of the County Law. The Council document also noted that ïnexplicablÿ 18-B lawyers are handling 50 percent of the indigent criminal cases in New York City, even though their mandate is to handle only multi-defendant cases where the Legal Aid Society had a conflict. In past years, the City Council had consistently added \$5.6 million to the \$54.7 million proposed for the Legal Aid Society by former Mayor Giuliani, bringing the total to just a shade over \$60 million. But this year for the first time, the Council is proposing shifting more than \$20 million in funds earmarked by the Mayor for 18-B lawyers to the Legal Aid Society, which would increase its total funding to \$80.4 million. That would reflect a jump in its current finding of about one-third. Meantime, the City Council proposed slashing the Mayor's allocation of \$62.8 million for 18-B lawyers by 66 percent, to \$21.4 million.\newline{}Question: By increasing current funding to the Legal Aid society by \$25.6 million, how much is the Council increasing their funding?\newline{}Choices: ['\$60 million', '\$62.8 million', 'One third', '\$54.7 million', '\$80.4 million']\newline{}Note: 1. there can be multiple correct answers. 2. each line contains one answer. 3. If no correct answer, reply 'none'.\newline{}Your answer: \bigstrut\\
\hline
PIQA  & Goal: how do you flood a room?\newline{}Choose the most sensible solution to achieve the goal. Choices: ['fill it with objects.', 'fill it with water.']\newline{}Answer: \bigstrut\\
\hline
ReCoRD & A father has admitted killing his 13-year-old son by giving him a morphine tablet when the boy complained that he was feeling ill. Kevin Morton gave his son Kye Backhouse an extremely strong painkiller, a court heard - a mistake which he says he will 'have to try and live with it for the rest of my life'. He could now face jail after pleading guilty to manslaughter over the teenager's death at Preston Crown Court. Tragedy: Kevin Morton, right, has admitted killing his son Kye Backhouse, left, by giving him morphine 'Happy-go-lucky' Kye was found dead at his home in Barrow-in-Furness, Cumbria in October last year.@highlight Morton gave Kye Backhouse a strong painkiller when he was ill@highlight teenager subsequently died and his father has admitted manslaughter@highlight, 49, faces jail when he is sentenced next month\newline{}Question: Death: @placeholder, 23, complained of feeling unwell before his father gave him the strong painkiller What is the @̈placeholder?̈\newline{}Answer: \bigstrut\\
\hline
RTE   & Premise: Euro Disney is one of the most popular theme parks of USA.\newline{}Hypothesis: Euro-Disney is an Entertainment Park.\newline{}Question: Determine whether the premise entails the hypothesis or not.\newline{}Choices: ['entailment', 'not\_entailment']\newline{}Answer: \bigstrut\\
\hline
WiC   & Sentence1: An early movie simply showed a long kiss by two actors of the contemporary stage.\newline{}Sentence2: We went out of town together by stage about ten or twelve miles.\newline{}Question: Does 'stage' have the same meaning in both sentences?\newline{}Choices: ['true', 'false']\newline{}Answer: \bigstrut\\
\hline
WinoGrande & Choose the most sensible text to replace the '\_' in the following sentence: Natalie was less religous than Patricia, therefore \_ attended church services more often on Sundays.\newline{}Choices: ['Natalie', 'Patricia']\newline{}Answer: \bigstrut\\
\hline
WSC   & The mothers of Arthur and Celeste have come to the town to fetch them. They are very happy to have them back, but they scold them just the same because they ran away.\newline{}Question: In the above text, does 'them' refer to 'mothers'?\newline{}Choices:['true', 'false']\newline{}Answer: \bigstrut\\
\hline
\caption{The template of each dataset.} 
\label{tab:templates}
\end{longtable}
\twocolumn

\end{document}